\documentclass[a4paper,twoside]{article}

\usepackage{epsfig}
\usepackage{subcaption}
\usepackage{calc}
\usepackage{graphicx}
\usepackage{amsmath}
\usepackage{subcaption}
\usepackage{amssymb}
\usepackage{hyperref}

\usepackage{booktabs}
\usepackage{comment}
\usepackage{caption} 
\usepackage{times}
\usepackage{epsfig}
\usepackage{adjustbox}

\usepackage{multirow}
\usepackage{times}
\usepackage{colortbl}
\usepackage{epsfig}
\usepackage{graphicx}
\usepackage{adjustbox}
\usepackage{amsmath}
\usepackage{amssymb}

\usepackage[utf8]{inputenc} % allow utf-8 input
\usepackage[T1]{fontenc}    % use 8-bit T1 fonts
\usepackage{url}            % simple URL typesetting
\usepackage{booktabs}       % professional-quality tables
\usepackage{amsfonts}       % blackboard math symbols
\usepackage{nicefrac}       % compact symbols for 1/2, etc.
\usepackage{bbm}            % blackboard bold fonts
\usepackage{enumitem}
\usepackage{float}
\usepackage{microtype}
\usepackage[ruled,vlined,linesnumbered]{algorithm2e}
\usepackage{enumitem}
\usepackage{amssymb}
\usepackage{amstext}
\usepackage{amsmath}
\usepackage{amsthm}
\usepackage{multicol}
\usepackage{pslatex}
\usepackage{apalike}
\usepackage{algorithm2e}
\usepackage[bottom]{footmisc}
\usepackage{SCITEPRESS}     % Please add other packages that you may need BEFORE the SCITEPRESS.sty package.
\makeatletter
\@namedef{ver@everyshi.sty}{}
\makeatother

\usepackage[dvipsnames,svgnames,x11names]{xcolor}
\usepackage{tikz}
\usetikzlibrary{arrows.meta,shapes,calc,matrix,fit,positioning,backgrounds,decorations.markings}
\usepackage{pgfplots}
\usepackage{pgfplotstable}
\pgfplotsset{compat=1.9}
\usepackage{xstring}

\usepgfplotslibrary{external}
\IfBeginWith*{\jobname}{fig/extern/}{\finalcopy}{}

\tikzstyle{every picture}+=[
	remember picture,
	every text node part/.style={align=center},
	every matrix/.append style={ampersand replacement=\&},
]
\tikzstyle{tight} = [inner sep=0pt,outer sep=0pt]
\tikzstyle{node}  = [draw,circle,tight,minimum size=12pt,anchor=center]
\tikzstyle{op}    = [draw,circle,tight]
\tikzstyle{dot}   = [fill,draw,circle,inner sep=1pt,outer sep=0]
\tikzstyle{pt}    = [fill,draw,circle,inner sep=1.5pt,outer sep=.2pt]
\tikzstyle{box}   = [draw,thick,rectangle,inner sep=3pt]
\tikzstyle{high}  = [black!60]
\tikzstyle{group} = [high,box,opacity=.5]
\tikzstyle{rectc} = [tight,transform shape]
\tikzstyle{rect}  = [rectc,anchor=south west]

\newcommand{\leg}[1]{\addlegendentry{#1}}

\tikzset{every mark/.append style={solid}}
\pgfplotsset{%smooth,
	grid=both, width=\columnwidth, try min ticks=5,
	every axis/.append style={font=\small},
	every axis plot/.append style={thick,mark=none,mark size=1.8,tension=0.18},
	legend cell align=left, legend style={fill opacity=0.8},
	xticklabel={\pgfmathprintnumber[assume math mode=true]{\tick}},
	yticklabel={\pgfmathprintnumber[assume math mode=true]{\tick}},
	nodes near coords math/.style={
		nodes near coords={\pgfmathprintnumber[assume math mode=true]{\pgfplotspointmeta}},
	},
}

\pgfplotsset{
	dash/.style={mark=o,dashed,opacity=0.6},
	dott/.style={mark=o,dotted,opacity=0.6},
	nolim/.style={enlargelimits=false},
	plain/.style={every axis plot/.append style={},nolim,grid=none},
}

\newcommand{\train}{\mathrm{train}}
\newcommand{\test}{\mathrm{test}}
\newcommand{\val}{\mathrm{val}}

\newcommand{\ee}{$\mathrm{E^2}$ }

\begin{document}

\title{Image edge enhancement for effective image classification}

\author{\authorname{Bu Tianhao\sup{1}, Michalis Lazarou\sup{2} and Tania Stathaki\sup{2}}
\affiliation{\sup{1}Glory Engineering \& Tech Co., LTD \\ \sup{2}Imperial College London}
% \email{\{bu, ml6414, s\_author\}@ic.ac.uk, t\_author@dc.mu.edu}
}

% !TEX root = ../paper.tex

\newcommand{\head}[1]{{\smallskip\noindent\textbf{#1}}}
\newcommand{\alert}[1]{{{#1}}}
\newcommand{\mich}[1]{\textcolor{blue}{#1}}

\newcommand{\sm}{\scriptsize}
\newcommand{\eq}[1]{(\ref{eq:#1})}

\newcommand{\Th}[1]{\textsc{#1}}
\newcommand{\mr}[2]{\multirow{#1}{*}{#2}}
\newcommand{\mc}[2]{\multicolumn{#1}{c}{#2}}
\newcommand{\tb}[1]{\textbf{#1}}
\newcommand{\ch}{\checkmark}

\newcommand{\red}[1]{{\color{red}{#1}}}
\newcommand{\blue}[1]{{\color{blue}{#1}}}
\newcommand{\green}[1]{{\color{green}{#1}}}
\newcommand{\gray}[1]{{\color{gray}{#1}}}

\newcommand{\citeme}[1]{\red{[XX]}}
\newcommand{\refme}[1]{\red{(XX)}}

\newcommand{\fig}[2][1]{\includegraphics[width=#1\columnwidth]{fig/#2}}
\newcommand{\figh}[2][1]{\includegraphics[height=#1\columnwidth]{fig/#2}}

%--------------------------------------------------------------------

\newcommand{\tran}{^\top}
\newcommand{\mtran}{^{-\top}}
\newcommand{\zcol}{\mathbf{0}}
\newcommand{\zrow}{\zcol\tran}

\newcommand{\ind}{\mathbbm{1}}
\newcommand{\expect}{\mathbb{E}}
\newcommand{\nat}{\mathbb{N}}
\newcommand{\zahl}{\mathbb{Z}}
\newcommand{\real}{\mathbb{R}}
\newcommand{\proj}{\mathbb{P}}
\newcommand{\prob}{\mathbf{Pr}}
\newcommand{\normal}{\mathcal{N}}

\newcommand{\mif}{\textrm{if}\ }
\newcommand{\other}{\textrm{otherwise}}
\newcommand{\minimize}{\textrm{minimize}\ }
\newcommand{\maximize}{\textrm{maximize}\ }
\newcommand{\st}{\textrm{subject\ to}\ }

\newcommand{\id}{\operatorname{id}}
\newcommand{\const}{\operatorname{const}}
\newcommand{\sgn}{\operatorname{sgn}}
\newcommand{\var}{\operatorname{Var}}
\newcommand{\mean}{\operatorname{mean}}
\newcommand{\trace}{\operatorname{tr}}
\newcommand{\diag}{\operatorname{diag}}
\newcommand{\vect}{\operatorname{vec}}
\newcommand{\cov}{\operatorname{cov}}
\newcommand{\sign}{\operatorname{sign}}
\newcommand{\prj}{\operatorname{proj}}

\newcommand{\softmax}{\operatorname{softmax}}
\newcommand{\clip}{\operatorname{clip}}

\newcommand{\defn}{\mathrel{:=}}
\newcommand{\peq}{\mathrel{+\!=}}
\newcommand{\meq}{\mathrel{-\!=}}

\newcommand{\floor}[1]{\left\lfloor{#1}\right\rfloor}
\newcommand{\ceil}[1]{\left\lceil{#1}\right\rceil}
\newcommand{\inner}[1]{\left\langle{#1}\right\rangle}
\newcommand{\norm}[1]{\left\|{#1}\right\|}
\newcommand{\abs}[1]{\left|{#1}\right|}
\newcommand{\frob}[1]{\norm{#1}_F}
\newcommand{\card}[1]{\left|{#1}\right|\xspace}
\newcommand{\diff}{\mathrm{d}}
\newcommand{\der}[3][]{\frac{d^{#1}#2}{d#3^{#1}}}
\newcommand{\pder}[3][]{\frac{\partial^{#1}{#2}}{\partial{#3^{#1}}}}
\newcommand{\ipder}[3][]{\partial^{#1}{#2}/\partial{#3^{#1}}}
\newcommand{\dder}[3]{\frac{\partial^2{#1}}{\partial{#2}\partial{#3}}}

\newcommand{\wb}[1]{\overline{#1}}
\newcommand{\wt}[1]{\widetilde{#1}}

\def\xssp{\hspace{1pt}}
\def\ssp{\hspace{3pt}}
\def\msp{\hspace{5pt}}
\def\lsp{\hspace{12pt}}

\newcommand{\cA}{\mathcal{A}}
\newcommand{\cB}{\mathcal{B}}
\newcommand{\cC}{\mathcal{C}}
\newcommand{\cD}{\mathcal{D}}
\newcommand{\cE}{\mathcal{E}}
\newcommand{\cF}{\mathcal{F}}
\newcommand{\cG}{\mathcal{G}}
\newcommand{\cH}{\mathcal{H}}
\newcommand{\cI}{\mathcal{I}}
\newcommand{\cJ}{\mathcal{J}}
\newcommand{\cK}{\mathcal{K}}
\newcommand{\cL}{\mathcal{L}}
\newcommand{\cM}{\mathcal{M}}
\newcommand{\cN}{\mathcal{N}}
\newcommand{\cO}{\mathcal{O}}
\newcommand{\cP}{\mathcal{P}}
\newcommand{\cQ}{\mathcal{Q}}
\newcommand{\cR}{\mathcal{R}}
\newcommand{\cS}{\mathcal{S}}
\newcommand{\cT}{\mathcal{T}}
\newcommand{\cU}{\mathcal{U}}
\newcommand{\cV}{\mathcal{V}}
\newcommand{\cW}{\mathcal{W}}
\newcommand{\cX}{\mathcal{X}}
\newcommand{\cY}{\mathcal{Y}}
\newcommand{\cZ}{\mathcal{Z}}
\newcommand{\cPi}{\mathcal{\pi}}

\newcommand{\vA}{\mathbf{A}}
\newcommand{\vB}{\mathbf{B}}
\newcommand{\vC}{\mathbf{C}}
\newcommand{\vD}{\mathbf{D}}
\newcommand{\vE}{\mathbf{E}}
\newcommand{\vF}{\mathbf{F}}
\newcommand{\vG}{\mathbf{G}}
\newcommand{\vH}{\mathbf{H}}
\newcommand{\vI}{\mathbf{I}}
\newcommand{\vJ}{\mathbf{J}}
\newcommand{\vK}{\mathbf{K}}
\newcommand{\vL}{\mathbf{L}}
\newcommand{\vM}{\mathbf{M}}
\newcommand{\vN}{\mathbf{N}}
\newcommand{\vO}{\mathbf{O}}
\newcommand{\vP}{\mathbf{P}}
\newcommand{\vQ}{\mathbf{Q}}
\newcommand{\vR}{\mathbf{R}}
\newcommand{\vS}{\mathbf{S}}
\newcommand{\vT}{\mathbf{T}}
\newcommand{\vU}{\mathbf{U}}
\newcommand{\vV}{\mathbf{V}}
\newcommand{\vW}{\mathbf{W}}
\newcommand{\vX}{\mathbf{X}}
\newcommand{\vY}{\mathbf{Y}}
\newcommand{\vZ}{\mathbf{Z}}

\newcommand{\va}{\mathbf{a}}
\newcommand{\vb}{\mathbf{b}}
\newcommand{\vc}{\mathbf{c}}
\newcommand{\vd}{\mathbf{d}}
\newcommand{\ve}{\mathbf{e}}
\newcommand{\vf}{\mathbf{f}}
\newcommand{\vg}{\mathbf{g}}
\newcommand{\vh}{\mathbf{h}}
\newcommand{\vi}{\mathbf{i}}
\newcommand{\vj}{\mathbf{j}}
\newcommand{\vk}{\mathbf{k}}
\newcommand{\vl}{\mathbf{l}}
\newcommand{\vm}{\mathbf{m}}
\newcommand{\vn}{\mathbf{n}}
\newcommand{\vo}{\mathbf{o}}
\newcommand{\vp}{\mathbf{p}}
\newcommand{\vq}{\mathbf{q}}
\newcommand{\vr}{\mathbf{r}}
\newcommand{\Vs}{\mathbf{s}}
\newcommand{\vt}{\mathbf{t}}
\newcommand{\vu}{\mathbf{u}}
\newcommand{\vv}{\mathbf{v}}
\newcommand{\uu}{\mathbf{u}}
\newcommand{\cc}{\mathbf{c}}
\newcommand{\vw}{\mathbf{w}}
\newcommand{\vx}{\mathbf{x}}
\newcommand{\vy}{\mathbf{y}}
\newcommand{\vz}{\mathbf{z}}

\newcommand{\vone}{\mathbf{1}}
\newcommand{\vzero}{\mathbf{0}}

\newcommand{\valpha}{{\boldsymbol{\alpha}}}
\newcommand{\vbeta}{{\boldsymbol{\beta}}}
\newcommand{\vgamma}{{\boldsymbol{\gamma}}}
\newcommand{\vdelta}{{\boldsymbol{\delta}}}
\newcommand{\vepsilon}{{\boldsymbol{\epsilon}}}
\newcommand{\vzeta}{{\boldsymbol{\zeta}}}
\newcommand{\veta}{{\boldsymbol{\eta}}}
\newcommand{\vtheta}{{\boldsymbol{\theta}}}
\newcommand{\viota}{{\boldsymbol{\iota}}}
\newcommand{\vkappa}{{\boldsymbol{\kappa}}}
\newcommand{\vlambda}{{\boldsymbol{\lambda}}}
\newcommand{\vmu}{{\boldsymbol{\mu}}}
\newcommand{\vnu}{{\boldsymbol{\nu}}}
\newcommand{\vxi}{{\boldsymbol{\xi}}}
\newcommand{\vomikron}{{\boldsymbol{\omikron}}}
\newcommand{\vpi}{{\boldsymbol{\pi}}}
\newcommand{\vrho}{{\boldsymbol{\rho}}}
\newcommand{\vsigma}{{\boldsymbol{\sigma}}}
\newcommand{\vtau}{{\boldsymbol{\tau}}}
\newcommand{\vupsilon}{{\boldsymbol{\upsilon}}}
\newcommand{\vphi}{{\boldsymbol{\phi}}}
\newcommand{\vchi}{{\boldsymbol{\chi}}}
\newcommand{\vpsi}{{\boldsymbol{\psi}}}
\newcommand{\vomega}{{\boldsymbol{\omega}}}

\newcommand{\rLambda}{\mathrm{\Lambda}}
\newcommand{\rSigma}{\mathrm{\Sigma}}

\newcommand{\vLambda}{\bm{\rLambda}}
\newcommand{\vSigma}{\bm{\rSigma}}

% big cdot
\makeatletter
\newcommand*\bdot{\mathpalette\bdot@{.7}}
\newcommand*\bdot@[2]{\mathbin{\vcenter{\hbox{\scalebox{#2}{$\m@th#1\bullet$}}}}}
\makeatother

%--------------------------------------------------------------------
% Add a period to the end of an abbreviation unless there's one
% already, then \xspace.
\makeatletter
\DeclareRobustCommand\onedot{\futurelet\@let@token\@onedot}
\def\@onedot{\ifx\@let@token.\else.\null\fi\xspace}

\def\eg{\emph{e.g}\onedot} \def\Eg{\emph{E.g}\onedot}
\def\ie{\emph{i.e}\onedot} \def\Ie{\emph{I.e}\onedot}
\def\cf{\emph{cf}\onedot} \def\Cf{\emph{Cf}\onedot}
\def\etc{\emph{etc}\onedot} \def\vs{\emph{vs}\onedot}
\def\wrt{w.r.t\onedot} \def\dof{d.o.f\onedot} \def\aka{a.k.a\onedot}
\def\etal{\emph{et al}\onedot}
\makeatother

\abstract{Image classification has been a popular task due to its feasibility in real-world applications. Training neural networks by feeding them RGB images has demonstrated success over it.  Nevertheless, improving the classification accuracy and computational efficiency of this process continues to present challenges that researchers are actively addressing. A widely popular embraced method to improve the classification performance of neural networks is to incorporate data augmentations during the training process. Data augmentations are simple transformations that create slightly modified versions of the training data, and can be very effective in training neural networks to mitigate overfitting and improve their accuracy performance. 
 In this study, we draw inspiration from high-boost image filtering and propose an edge enhancement-based method as means to enhance both accuracy and training speed of neural networks. Specifically, our approach involves extracting  high frequency features, such as edges, from images within the available dataset and fusing them with the original images, to generate new, enriched images. Our comprehensive experiments, conducted on two distinct datasets—CIFAR10 and CALTECH101, and three different network architectures—ResNet-18 ,LeNet-5 and CNN-9—demonstrate the effectiveness of our proposed method.
}

\onecolumn \maketitle \normalsize \setcounter{footnote}{0} \vfill
\section{\uppercase{Introduction}}
\label{sec:introduction}

Deep learning has undeniably been at the forefront of technological advancement over the past decade, with a transformative influence, particularly in the field of computer vision~\cite{he2016deep}. However, training neural networks effectively for computer vision tasks poses many limitations, because of the reliance on substantial amount of data to avoid overfitting\cite{zhang2019overfitting}.

Data augmentation is a widely recognized technique for mitigating model overfitting and enhancing generalization, effectively addressing the issue of insufficient sample diversity\cite{shao2023fads}. A common practice within the realm of data augmentation is to apply colour space and$/$or other geometric transformations to the original images in order to increase the size and diversity of the training dataset. Some notable works of data augmentation techniques include image mix-up approaches, such as the creation of new images by directly merging two images from distinct classes  \cite{zhang2017mixup}, replacing specific regions of an image with patches from other  images \cite{yun2019cutmix} and various other inventive variations \cite{liu2022automix}, \cite{yin2021batchmixup}, \cite{kim2020puzzle}.  Other approaches utilize the frequency domain information \cite{chen2021few}, \cite{mukai2022improving} with a focus on generating new sets of images from diverse sources. 

In our work, we present a novel data augmentation method that transforms the original images by merging them with their corresponding extracted edges,  that we will refer to as \emph{Edge Enhancement} ($\mathrm{E}^2$). It operates on the same principle as high boost image filtering, a technique that accentuates the edges of an image without completely eliminating the background. The proposed \ee method extracts the edges (high frequency components) from each of the RGB channels of the original images and subsequently integrates these extracted features with the original image to yield the edge-enhanced images. \ee adheres to the fundamental concept of introducing subtle modifications by utilizing these extracted features to generate new images, all without relying on complex algorithms, thus ensuring efficient preprocessing.

%----------------------------------------------------------------------------------------
\begin{figure*}[ht]
  \centering
\includegraphics[width=\textwidth]{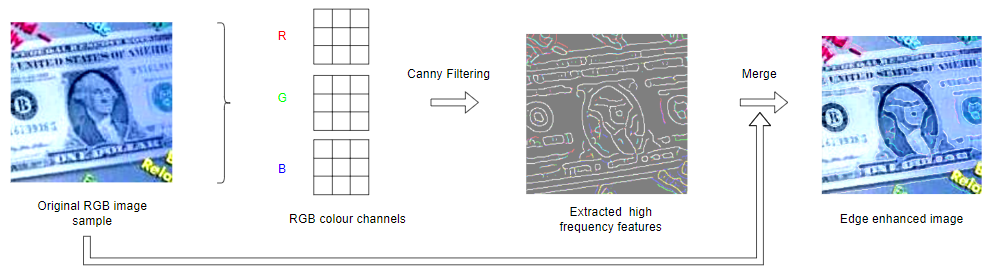}  
\caption{Edge enhancement process. Leftmost: original image. Mid left: corresponding R,G,B channels of the original labelled data. Mid right:corresponding extracted edges Rightmost: edge enhanced image.}
\label{fig:ee}
 \end{figure*}
%-------------------------------------------------------------------------------------------

The main contributions of our work are summarized as follows:
\begin{enumerate}

\item We propose a novel data augmentation method that enhances the semantic information of every image in the training dataset which in turn improves the training of neural networks. 

\item Our method has undergone rigorous evaluation on two distinct datasets, CIFAR10 and CALTECH101, employing neural network models including CNN-9, LeNet-5 and RESNET-18. These experiments have consistently showcased substantial improvements in classification accuracy, accompanied by a reduced requirement for training epochs to achieve optimal classification accuracy. 
\end{enumerate}

\section{\uppercase{Background}}
\label{sec:related}

\subsection{Data augmentation}
Data augmentation is a technique that expands dataset sizes by introducing subtle alterations.

The simplest data augmentation strategies are through applying geometric transformations as rotation and flipping \cite{chlap2021review}. Our research aligns with this principle of generating new labeled data by modifying the original dataset. However, it deviates from the conventional methods of rotation and flipping. Instead, we propose a novel approach that prioritizes the enhancement of semantic information. This innovative method draws inspiration from the concept of high-boost filtering, which will be elaborated upon in the subsequent section.

\subsection{High boost filtering}
High-boost filtering is a well-established image processing technique, representing a variation of high-pass filtering. In contrast to high-pass filtering, high-boost filtering aims to accentuate an image's high-frequency information while preserving the background.
Let's denote an image undergoing this process as I(u,v) \cite{srivastava2009pde}, where u,v correspond to spatial coordinates. High boost filtering can be formally defined as follows: 
\begin{equation}\label{eq1}
    I_{boost}(u,v)=A\cdot I(u,v)-I_{low}(u,v)
\end{equation}

where $I_{boost}(u,v)$ is the high boosted image pixel at location u, v, and A is a scaling factor greater than 1 (common choices are 1.15, 1.2) and $I_{low}(u,v)$ is the low pass filtered image pixel. Respectively, the high pass filtering can be presented in a similar way as,  
\begin{equation}\label{eq2}
    I_{high}(u,v)=I(u,v)-I_{low}(u,v)
\end{equation}
where $I_{high}(u,v)$ is the high pass filtered image pixel at location u, v. 
Our method employs the concept of high-boost filtering, where we combine the high-pass filtered image with the original version, with A = 2. This selection for A can be viewed as a particular instance of high-boost filtering.

%----------------------------------------------------------------------------------------
\begin{figure*}[ht]
  \centering
\includegraphics[width=0.62\textwidth, height = 0.42\textwidth]{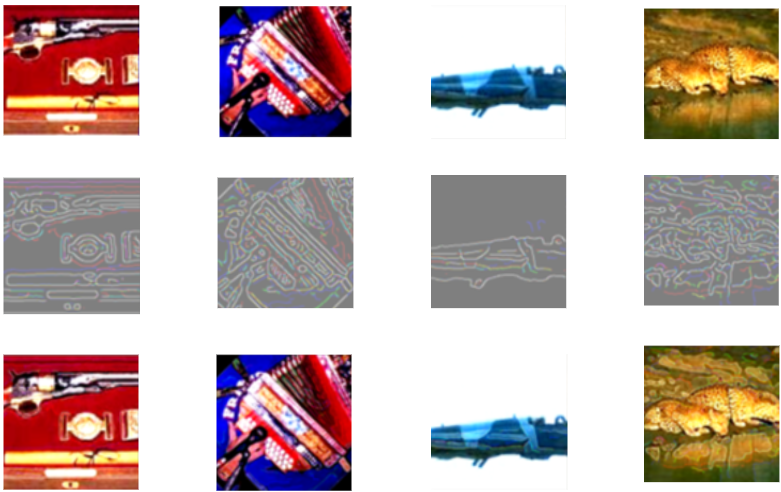}  
\caption{A batch of 4 images. Top row: original  data. Mid row: corresponding extracted edges. Bottom row: corresponding edge enhanced data  }
  \label{fig:batch4}
 \end{figure*}
%-------------------------------------------------------------------------------------------
\subsection{Canny edge detection}

Canny edge detection algorithm was invented by John F Canny in 1986 \cite{canny_original}. The procedure of canny edge detection can be shown as follows:
1) Noise suppression within the image is accomplished by employing a Gaussian filter (a low-pass filter) \cite{deng1993adaptive}. The kernel for the 2-D Gaussian filter can be represented as: 
\begin{equation}\label{eq3}
    G\left(u,v\right)=\frac{1}{2\pi\sigma^2}e^{-\frac{u^2+v^2}{2\sigma^2}}
\end{equation}
The denoised result is achieved by convolving the kernel with the image, where $\sigma^2$ represents the filter variance.
2) Determine the edge strength $ \mathrm{\nabla G}$ and the four orientations $\theta$ (horizontal, vertical and the two diagonals), expressed as
\begin{equation}\label{eq4}
    \mathrm{\nabla G}={(\frac{\partial G}{\partial u}+\frac{\partial G}{\partial v})}^{0.5}
\end{equation}

\begin{equation}\label{eq5}
    \theta=\tan^{-1}{(\frac{\partial G}{\partial v}/\frac{\partial G}{\partial u})}
\end{equation}
where $\frac{\partial G}{\partial u},\frac{\partial G}{\partial v} $ are the first order derivatives (gradients) in the horizontal and vertical directions. 
3) Non-maximum suppression, which utilizes the edge strength to suppress non-edge pixels. 
4) Double thresholding, a method which involves establishing both high and low edge strength thresholds, which are employed to categorize the strength of edge pixels into three levels: strong, weak, and non-edges. Additionally, non-edge pixels are subsequently suppressed.  
5) Hysteresis edge tracking represents the final step, employing blob analysis to further handle weak edges. Those weak edges which lack a strong edge neighbor connection are identified as potential noise or color variation-induced edges and subsequently removed.
In our work we use Canny edge detection as a core technique to extract the edge information in each colour channel from the dataset of interest.

\section{\uppercase{methodology}}
\subsection{Problem definition}
Let us define a labeled image dataset $ D\ =\ (\vx_i,\vy_i)$, where $\vx_i $ represents the $i^{th}$  image and $\vy_i$ represents the corresponding class label of image $\vx_i $.This dataset comprises the images, each containing three RGB channels. The dataset is partitioned into three splits: the training set split, ${D_{\train}=\ ({\vx}_i,\vy_i})$, the validation set split,  $ {D_{\val}=\ ({\vx}_i,\vy_i})$ and testing set split, ${D_{\test}=\ ({\vx}_i,\vy_i}).$ We use $D_{\train}$ to train a neural network, that consists of a backbone $f_{\theta}$ and a classifier $g_{\phi}$ (last layer of the network). Additionally, we denote the edge features extracted from the training data as $\vv_i$ ,  this is merged with $\vx_i$ to create the edge enhanced data $\ve_i$. 

The validation set $D_{\val}$ is used in order to save the model with the highest validation accuracy. Finally, we use $D_{\test}\ $ to calculate the test set classification accuracy.

\subsection{Edge enhancement}

Our \ee pre-processing can be illustrated in \autoref{fig:ee}, where the leftmost image is the original data. After applying Canny filtering the extracted edges are shown and the rightmost image is the edge enhanced result.  We first apply normalization to the sample images, and then resize the images to ensure all images have exactly the same resolution, also maintaining consistent image size across the datasets. Canny filters are employed to the every image at each of its RGB channels to obtain the extracted edge information. The extracted edge information is then combined with their respective original versions, resulting in a transformed image set. Each image in this set is labeled according to the original class labels. 
%----------------------------------------------------------------------------------------
\begin{figure*}[h]
  \centering
\includegraphics[width=\textwidth, height = 0.22\textwidth]{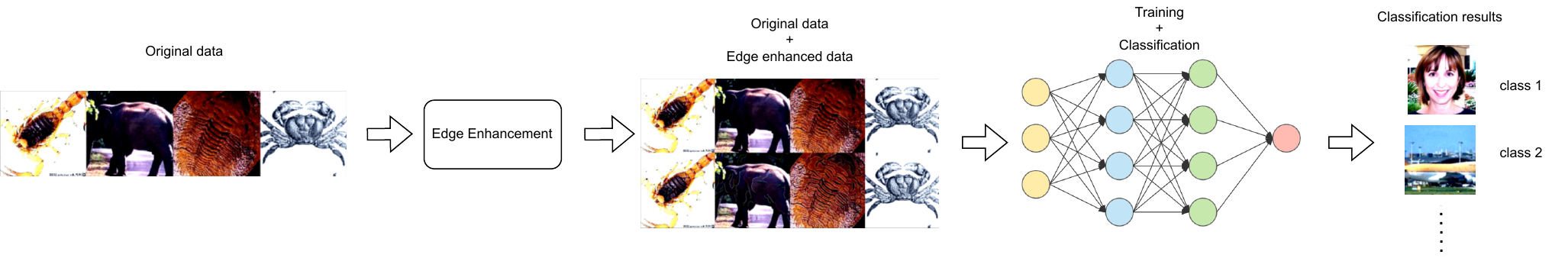}  
\caption{The flowchart of the edge enhancement method for model training. i) Original data is edge enhanced to the transformed data. ii) The original and transformed data is concatenated then feed into the network. iii) Model training. iv) Classification results obtained through test dataset.}
  \label{fig:example1}
 \end{figure*}
%-------------------------------------------------------------------------------------------

\subsection{Training phase}

In each batch, we apply edge enhancement to a set of training images, as depicted in \autoref{fig:batch4}. In this representation, the top four images represent the original data, while the bottom four belong to the edge-enhanced images, corresponding to the edge-enhanced dataset ${\widetilde{D}}=\ {{\widetilde{\vx}},{\widetilde{\vy}}}$, where $\widetilde{\vx}$ is the edge-enhanced image and $\widetilde{\vy}$ is the corresponding class label of image $\widetilde{\vx}$. These edge-enhanced images are combined with the original training set images to create the input data for the network model. Consequently, each batch sample comprises a total of eight images, contributing to the generation of classification results. 

During the forward propagation, we extract the features $\vv_i$ by inputting every image $\vx_i$ into the Canny edge detection function, denoted as $Canny()$. These features can be expressed as follows:
\begin{equation}
    \widetilde{\vx_i} \defn Canny(\vx_i) + \vx_i
\end{equation}

 During the training phase, the edge-enhanced and original images are both used to train the network by concatenating them together before feeding them to the network. Therefore, every batch size now contains both the edge enhanced images, $\{\widetilde{\vx_i}\}_{i=1}^{b}$, and the original images, $\{\vx_{i}\}_{i=1}^{b}$, where $b$ represents the batch size. After, the concatenation we have $\{\vu_{i}\}_{i=1}^{2b}$ because the size of the new batch size is $2b$. We feed every image $u_i$ from the new batch to network $f_{\theta}$ to obtain the feature vector $\vz_{i}$ that corresponds to image $\vu_i$, expressed as 

\begin{equation}
\begin{split}
    % \vz_{x_i} & \defn f_{\theta}(\vx_i) \\
    \vz_{i}  & \defn  f_{\theta}(\vu_i)
\end{split}
\end{equation}
We pass $\vz_{i} $ to $g_{\phi}$  so that 
%{{\vt_{y_i},\vt_{y_i}}\}
\begin{equation}
    \vc_i \defn g_{\phi}(\vz_{i})
\label{eq:logits}
\end{equation}
where $\vc_i$ is the vector of logits of image $i$. We obtain the probability of each class $\hat{p_i}$ by using the softmax activation function

\begin{equation}
    \hat{p_i} \defn \frac{\exp(\vc_i)}{\sum_{i=1}^N \exp(\vc_{i})},
\label{eq:softmax}
\end{equation}
where $N$ is the total number of classes. With the aid of equation \eq{softmax}, we calculate the cross-entropy loss $L_{ce}$ as 
\begin{equation}
   L_{ce} \defn -\displaystyle \sum_{i \in N}\overline{y_i}\log(\hat{p_i}),
\label{eq:ce}
\end{equation}
where $\overline{y_i}$ represents the one hot encoded version of $\vy_i$. Back propagation is then performed to calculate the gradient of $L_{ce}$ with respect to the network weight. 

Algorithm \ref{alg:train} summarizes the training procedure of \ee, and the corresponding flowchart is illustrated in \autoref{fig:example1}

\subsection{Inference stage}

During the inference stage, we calculate the probability vector, $\hat{p_i}$, for each image $x_i$ in the test dataset, $D_{\test}$ by embedding image $\vx_i$ using the functions $f_{\theta}$, $g_{\phi}$ and equation \eq{softmax}. To predict the class label $\hat{\vy}_{i}$ of image $\vx_i$ we employ the following formula:
\begin{equation}
	 \hat{\vy}_{i} \defn \arg\max_{j \in [N]} \hat{p}_{ij}
\label{eq:argmax}
\end{equation}
where $N$ is the number of possible classes and the result corresponds to the position of the element with the highest class probability in the class probability vector $p_i$.

We calculate the test dataset accuracy using the following formula:
\begin{equation}
    \vS_c \defn Acc(\vy_i,\hat{\vy}_i)
\end{equation}
where $\vS_c$ is the overall classification accuracy, and $Acc$ is the accuracy function that sums up the numbers of correct comparison results between the prediction and true class labels, we denote the sum as $N_{correct}$. To scale them into percentage, the results are divided by total number of images and multiplied by a hundred, we express it as $\frac{N_{correct}}{N_{total}}\times 100$.

\begin{algorithm}[!h]
 \caption{Edge enhancement training procedure}
 \label{alg:train}

 \SetKwInOut{Input}{input}
 \SetKwInOut{Output}{output}
 \Input{ training dataset ${D_{\train}=\ {\vx_i},\vy_i}$}
 
%\Input{ edge enhanced dataset ${\widetilde{D}}_t=\ {{\widetilde{\vx}}_i^t,{\widetilde{\vy}}_i^t}$}

\Output{ trained neural network model}
 {
  batch size: 4; \\
  % \BlankLine
\If{$D_{\train}\ \in\ \ $ CALTECH101}
 {$D_{\train}\ $= Resize($D_{\train}$, 256 x 256)\\ 
 $D_{\train}\ $= CenterCrop($D_{\train}$, 224 x 224)  }
 $D_{\train}$ = Normalization($D_{\train}$) \\
 % \BlankLine
 \While{epoch not finished}
 { \For{ i <= $D_{\train}$.size() }{
  {feature = Canny($\vx_i$,sigma = 3),
  ${ \widetilde{\vx}}_i $ = add($\vx_i$,feature),
 Inputs = concat($\vx_i$ , ${\widetilde{\vx}}_i$) 
}
 }
  Model = Model.train(Inputs, $D_{\val}$)
 }}
\end{algorithm}

\section{\uppercase{experiments}}
\label{sec:experiments}

\pgfplotstableread{
        epochs        edge-enhancement       original
        1              57.82                 55.06   
        % 2              63.6                  60.48  
        % 3              68.66                 65.84
        4              67.56                 61.92
        % 5              70.42                 68.68
        % 6              69.22                 70.98
        7              67.34                 70.38
        % 8              72.62                 71.42
        % 9              75.34                 71.42
        10             73.08                 70.14
        % 11             73.86                 68.68
        % 12             74.72                 71.16
        13             70.26                 66.22
        % 14             72.56                 70.04
        % 15             72.84                 69.58
        16             74.5                  70.48
        % 17             74.3                  72.34
        % 18             74.16                 72.74
        19             75.06                 72.28
        % 20             73.26                 72.42
        % 21             75.72                 69.08
        22             73.32                 69.38
        % 23             73.24                 70.48
        % 24             72.4                  52.56
        25             76.68                 68.16
        % 26             74.74                 68.26
        % 27             75.58                 69.9
        28             73.86                 71.24
        % 29             75.5                  69.96
        % 30             76.24                 71.84
        31             74.5                  71.2
        % 32             76.58                 70.26
        % 33             76.22                 72.22
        34             72.66                 71.98
        % 35             75.84                 71.88
        % 36             76.34                 68.94
        37             76.32                 73
        % 38             75.2                  72.76
        % 39             75.44                 71.78
        40             75.26                 72.4
        % 41             76.22                 70.82
        % 42             76.18                 73.5
        43             74.62                 73.68
        % 44             73.92                 71.66
        % 45             73.6                  71.8
        46             74.1                  72.56
        % 47             75.2                  67.12
        % 48             75.04                 69.88
        49             76.7                  70.48
        50             76.1                  66.92
}{\cnncifar}

\pgfplotstableread{
        epochs       edge-enhancement    original 
        1              40.5                 38.16   
        % 2              42.42                41.68     
        % 3              44.22                46.68
        4              45.02                46.28  
        % 5              49.78                44.32
        % 6              49.52                45.76
        7              50.00                44.46
        % 8              48.74                45.9
        % 9              46.82                48.32
        10             50.94                47.64
        % 11             44.90                44.14
        % 12             50.82                48.08
        13             50.66                45.6
        % 14             53.26                51.14
        % 15             47.16                46.14
        16             52.52                51.38
        % 17             52.5                 49.4
        % 18             49.46                49.7
        19             50.42                50.1
        % 20             51.02                47.3
        % 21             51.42                46.1
        22             52.4                 48.1
        % 23             51.48                50.6
        % 24             51.46                52.3
        25             50.68                49.5
        % 26             54.06                52
        % 27             50.94                49.3
        28             50.18                50.1
        % 29             54.16                51.5
        % 30             55                   45.5
        31             52.9                 45.9
        % 32             54.16                49.6
        % 33             55.74                48.5
        34             52.34                46.6
        % 35             51.74                51.3
        % 36             50.88                52.7
        37             52.64                48.18
        % 38             52.48                49
        % 39             50.64                50.12
        40             54.28                51
        % 41             53.28                50.1
        % 42             50.56                51.9
        43             52.28                44.02
        % 44             50.72                50.9
        % 45             52.12                53.4
        46             55.42                46.76
        % 47             48.04                45.6
        % 48             52.92                50.4
        49             49.6                 51.5
        50             53.54                48.38
}{\lenetcifarten}

\pgfplotstableread{
        epochs       edge-enhancement    original 
        1                 19.862            21.416
        % 2                 22.107            24.007
        % 3                 25.043            28.67 
        4                 30.743            30.397
        % 5                 32.988            36.097
        % 6                 37.478            34.197
        7                 37.478            38.687  
        % 8                 44.732            43.178
        % 9                 46.459            48.359  
        10                48.532            50.777   
        % 11                49.224            51.295 
        % 12                51.641            54.922  
        13                55.613            57.513  
        % 14                56.304            55.268  
        % 15                56.822            55.959  
        16                56.822            60.794 
        % 17                60.622            62.694 
        % 18                65.803            62.522 
        19                64.076            63.731
        % 20                59.24             63.558  
        % 21                62.522            62.176
        22                65.976            61.485
        % 23                61.14             62.522
        % 24                63.04             63.903 
        25                64.421            59.24     
        % 26                64.249            60.449
        % 27                60.967            66.839  
        28                63.04             63.903  
        % 29                63.731            65.803 
        % 30                65.285            64.421 
        31                59.585            63.731    
        % 32                64.94             63.212  
        % 33                63.903            65.803  
        34                65.803            64.594 
        % 35                64.594            66.148  
        % 36                67.012            63.385 
        37                68.048            64.249 
        % 38                61.485            63.731  
        % 39                61.658            65.975
        40                57.513            60.276  
        % 41                62.867            58.722  
        % 42                67.184            65.112 
        43                59.758            61.485  
        % 44                63.731            63.731  
        % 45                60.967            64.076  
        46                65.285            60.276
        % 47                63.731            57.858  
        % 48                63.903            62.867  
        49                60.276            63.558 
        % 50                63.04             65.803
        % 51                64.028            63.558
        52                65.823            62.003
        % 53                63.131            62.522
        % 54                67.17             66.667
        56                69.775            66.839
        % 57                66.67             65.63
        % 58                67.703            58.549
        59                68.739            58.377
        % 60                57.686            67.53
        % 61                55.786            64.94
        62                61.485            67.012   
        % 63                65.112            62.522
        % 64                65.112            60.967
        65                63.558            60.276
        % 66                62.867            57.168
        % 67                60.449            62.349
        68                60.967            69.085
        % 69                68.221            66.494
        % 70                69.775            64.594
        71                65.63             64.94
        % 72                63.731            63.558
        % 73                59.413            63.212
        74                61.313            61.658
        % 75                59.067            65.976
        % 76                59.24             64.767
        77                66.149            65.112
        % 78                65.976            64.249
        % 79                67.358            65.285
        80                68.048            61.831
        % 81                67.358            65.285
        % 82                64.076            60.276
        83                59.585            65.112
        % 84                59.585            65.803
        % 85                62.522            66.149
        86                65.285            69.275
        % 87                63.212            66.321
        % 88                68.739            57.858
        89                67.358            55.959
        % 90                67.185            60.276
        % 91                67/185            66.494
        92                53.541            67.012
        % 93                57.34             69.085
        % 94                63.212            65.976
        95                61.658            65.285
        % 96                67.185            59.067
        % 97                66.667            55.786
        98                67.185            59.057
        % 99                62.867            63.731
        100               57.686            60.449

        }{\caltech}
\subsection{Setup}
\paragraph{Datasets}
In our experiments two datasets are used: 1) CIFAR10, which is one of the most popular data sets in image classification research \cite{recht2018cifar}. It contains a total of 60000 color images of 10 classes,  each with dimensions of 32 by 32 pixels. In our experiment we use 45000 images for training, 5000 images for validation and 10000  for testing.

2) CALTECH101, considered as one of the challenging datasets \cite{bansal2021transfer}, CALTECH101 features higher resolution images in comparison to CIFAR10, as depicted in \autoref{fig:ee},\autoref{fig:batch4}, and \autoref{fig:example1}. This dataset contains 101 classes and includes 8,677 images.  For consistency in feeding images into the networks, we resized each image to 256 by 256 pixels and then center-cropped them to 224 by 224 pixels. Out of these images, 5,205 were used in the training set, 579 for validation, and 2,893 for testing.

\iffalse 
%------------------------------------------------------------------------------
\begin{table}[ht]
\small
\centering
\caption{Network and data set combinations with their corresponding learning rates.}

\setlength\tabcolsep{5pt}
\begin{tabular}{lc}
\toprule
 \Th{Network}                                              & \Th{learning rate }                  \\ \midrule
                                       \mc{2}{\Th{CIFAR10}}  \\ \midrule

%\midrule

LeNet-5                & 0.01  \\
CNN-9                         &0.01    \\

\midrule
                                       \mc{2}{\Th{CALTECH101}}  \\ \midrule
ResNet-18            & 0.01         \\                              
\bottomrule
\end{tabular}
\vspace{0pt}

\label{tab:lr}
\end{table}
%------------------------------------------------------------------------------
 \fi
\paragraph{Implementation details}

Our implementation is based in Python. We utilized the skimage package for the Canny edge detector and Pytorch for training our neural networks.

\paragraph{Networks}
In our work, we used three different neural network architectures tailored to the two datasets to investigate the robustness of our method with different networks and datasets. 1) LeNet-5: This convolutional neural network (CNN) originally proposed by LeCun in 1998\cite{lecun1998gradient}, was initially designed for hand written character classification. Given its compatibility with 32 by 32 pixel images and suitability for datasets with small class sizes (10 in CIFAR10),  we selected this network for the CIFAR10 dataset, due to its nature of accepting 32 by 32 images and suitable for small class sizes (10 for CIFAR10). 2) CNN-9: A standard neural network model with six convolutional layers and three linear layers, again suitable for 32 by 32 images. 3) ResNet-18: This network, proposed by \cite{he2016deep} in 2016. It is designed for larger images of size 224 by 224 pixels. We used this network for the high-resolution CALTECH101 dataset after preprocessing the data to meet the network's input requirements.

\paragraph{Hyperparameters}
In the experiment, the batch size is fixed at 4, meaning 4 labeled images were enhanced with edge information simultaneously. We set the standard deviation $\sigma$ of the Canny edge detector to 3 and trained the CIFAR10 dataset with LeNet-5 and CNN-9 for 50 epochs. For the CALTECH101 dataset, we trained ResNet-18 for 100 epochs. The loss function used is the cross-entropy loss due to its strong performance in multi-class classification tasks.\cite{zhou2019mpce}.\BlankLine
For weight optimization during network training, we used Stochastic Gradient Descent (SGD) with a momentum of 0.9, weight decay of 0.0005,  and  learning rate of 0.01. We use the validation set to calculate the validation accuracy and save the model with the highest validation accuracy through comparisons at each epoch.

%-------------------------------------------------------------------------------------
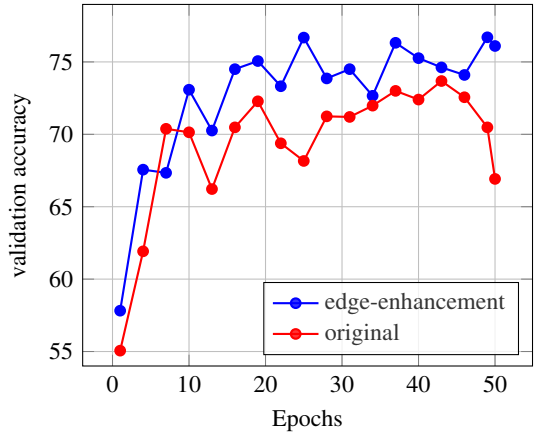
\begin{figure}
\begin{tikzpicture}
\begin{axis}[
    legend columns=1,
	width=\linewidth,
	height=0.85\linewidth,
	font=\small,
	ymin=54,
    % x dir=reverse,
	xlabel={Epochs},
	ylabel={validation accuracy},
	legend pos={south east},
]
% \addplot [black, domain=0:3.5] {25*x^0.2};
\addplot[blue, mark=*] table[x=epochs, y=edge-enhancement] \cnncifar; \leg{edge-enhancement}
\addplot[red, mark=*] table[x=epochs, y=original] \cnncifar; \leg{original}
\end{axis}
\end{tikzpicture}
\caption{\emph{CIFAR10 through CNN-9}}
\vspace{6pt}
\label{fig:alpha_fig}
\end{figure}
%------------------------------------------------------------------------------

\subsection{Ablation study}
\begin{figure}
\begin{tikzpicture}
\begin{axis}[
    legend columns=1,
	width=\linewidth,
	height=0.85\linewidth,
	font=\small,
	ymin=37,
    % x dir=reverse,
	xlabel={Epochs},
	ylabel={validation accuracy},
	legend pos={south east},
]
% \addplot [black, domain=0:3.5] {25*x^0.2};
\addplot[blue, mark=*] table[x=epochs, y=edge-enhancement] \lenetcifarten; \leg{edge-enhancement}
\addplot[red, mark=*] table[x=epochs, y=original]  \lenetcifarten; \leg{original}
\end{axis}
\end{tikzpicture}
\caption{\emph{CIFAR10 through LeNet-5}}
\vspace{6pt}
\label{fig_lenet}
\end{figure}
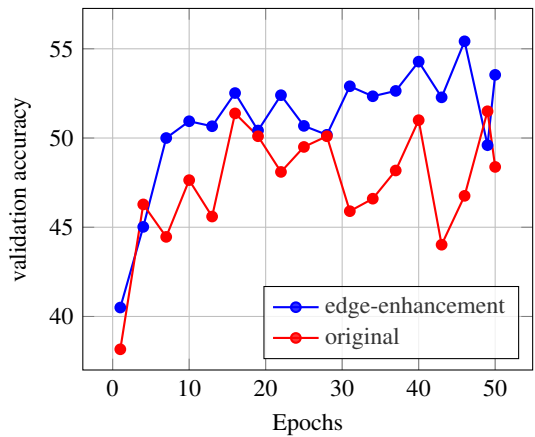

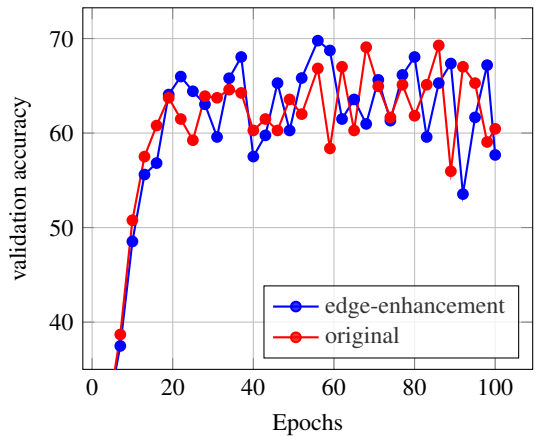
\begin{figure}
\begin{tikzpicture}
\begin{axis}[
    legend columns=1,
	width=\linewidth,
	height=0.85\linewidth,
	font=\small,
	ymin=35,
    % x dir=reverse,
	xlabel={Epochs},
	ylabel={validation accuracy},
	legend pos={south east},
]
% \addplot [black, domain=0:3.5] {25*x^0.2};
\addplot[blue, mark=*] table[x=epochs, y=edge-enhancement] \caltech; \leg{edge-enhancement}
\addplot[red, mark=*] table[x=epochs, y=original]  \caltech; \leg{original}
\end{axis}
\end{tikzpicture}
\caption{\emph{CALTECH101 through ResNet-18}}
\vspace{6pt}
\label{fig_resnet}
\end{figure}

In our experimental setup, for each training dataset and network architecture, we compared our edge enhancement method with the baseline method, which involved training on the original data without any modifications. We assessed this comparison through two aspects: 1) Validation Accuracy Variation: We monitored the validation accuracies for both methods throughout the training process, recording them at each epoch and presenting the results in graphical form. 2) Testing Accuracy: After identifying the models that achieved the best validation results, we tested them on the respective testing datasets to obtain optimal classification accuracies, which were then presented in a tabular format.  \BlankLine
1) CIFAR10: \autoref{fig:alpha_fig} and \autoref{fig_lenet} display the results obtained from both LeNet-5 and CNN-9 networks. In these figures, the edge enhancement method is represented by the blue lines, and it consistently outperforms the original method indicated by the red lines. The validation accuracy of the edge enhancement method surpasses that of the original at nearly every epoch. Under LeNet-5, the edge enhancement method achieves its highest accuracy at  55.74$\%$, differing from its lowest accuracy (40.05$\%$) by a substantial 15.69$\%$. In contrast, the original method exhibits a maximum-minimum difference of 15.24$\%$. Similarly, with CNN-9, the edge enhancement method shows a maximum-minimum difference of 18.86$\%$, compared to the original method's 18.62$\%$. These results clearly indicate that the edge enhancement method consistently provides greater accuracy improvement during the training process. Furthermore, our results reveal a noteworthy trend: for both network architectures within the CIFAR10 dataset, the validation accuracy variations are notably more stable and compact when using the edge enhancement method. CALTECH101: In \autoref{fig_resnet}, we present the results obtained through ResNet-18. Here, the edge enhancement method demonstrates a maximum-minimum difference of 44.73$\%$, while the original method yields a value of 40.61$\%$. Once again, these findings underscore the effectiveness of our method in achieving more substantial accuracy improvements.

\BlankLine 2) In \autoref{tab:soa-balanced}, we present the testing accuracy results, where \ee denotes the edge enhancement method, and 'epoch' indicates the epoch at which the optimal model was achieved. CIFAR10: For the CIFAR10 dataset employing LeNet-5, our method stands out with an impressive 5.7$\%$ increase in accuracy over the original approach. Moreover, it achieves this remarkable performance while converging 12 epochs faster to reach an optimal model. In the case of CNN-9 applied to CIFAR10, our method still shines, delivering a substantial 4.03$\%$ accuracy improvement. Notably, it accomplishes this while requiring 18 fewer epochs to converge. These outcomes underscore the significant enhancements our method brings to both network architectures when working with the CIFAR10 dataset.

CALTECH101: Through ResNet-18, our method demonstrated a 1.56$\%$ improvement over the original method, required 30 less epochs to converge.\BlankLine
To conclude, the experiment results are sufficient to show that the proposed edge enhancement method for neural network training provide noticeable classification accuracy improvement and it is more computationally efficient.

\subsection{Comparisons with transformation methods}
To further investigate the effectiveness of our edge enhancement method, we conducted a comprehensive evaluation by integrating it with various geometric transformation techniques.  Specifically, we assessed its performance in combination with transformations like random horizontal flipping and random image size cropping. Our objective was to compare the outcomes with a baseline approach that underwent the same corresponding transformations. We refer to the use of Random Cropping as C and Random Flipping as F.

In this experiment, we use the pytorch built-in transform functions and we split the comparisons within each dataset into three types: 1. exclusively applying random horizontal flipping. 2. solely implementing random image cropping . 3. simultaneously employing both random flipping and random cropping. The image classification accuracy results specified to each dataset section are compared, along with the results obtained without any geometric transformations, as shown in \autoref{tab:soa-balanced}. 
\paragraph{Random horizontal flip}
The "random horizontal flip" operation involves flipping the input data horizontally with a specified probability. In our experiment, we maintained a default probability of 50 percent for applying this random flipping to the input data.
\paragraph{Random image crop} 
The "random image crop" operation involves selecting a random location within the given input data and then resizing the cropped portion to a specified size. In our experiment, we adapted the resizing process based on the dataset being used: for CIFAR10, we resize the cropped data to 32 by 32 matching the original data size, suitable for LeNet-5 and CNN-9; for CALTECH101, since we use ResNet-18 for this data set, we again resize the cropped data to 224 by 224 fitting the requirement.
 \BlankLine

As the results shown in \autoref{tab:two}, when applied to the CIFAR10 dataset with network LeNet-5 our edge enhancement method provides significant accuracy improvements over the baseline under every transformation scenario. These improvements range from 3.16$\%$ to 3.97$\%$ when compared to the baseline. Notably, even when combined with random flipping, our method performs just 0.32$\%$  worse than our method without any transformations, as detailed in \autoref{tab:soa-balanced}. Turning our attention to the Caltech101 dataset with ResNet-18, \autoref{tab:three} showcases a similar trend. Our edge enhancement method consistently outperforms the original approach for all selected transformation combinations, with improvements ranging from 0.96$\%$ to 1.97$\%$. In this case, both edge enhancement with random cropping and edge enhancement with both cropping and flipping surpass the unmodified method (with no transformations) from  \autoref{tab:soa-balanced} in terms of performance. Furthermore, when considering the impact of these transformation techniques, our results affirm the claim that edge enhancement substantially enhances image classification accuracy. Intriguingly, for both datasets using LeNet-5 and ResNet-18, the original method with random cropping and flipping demonstrates a faster convergence to optimal performance compared to all other results.

%-------------------------------------------------------------------------------------
%------------------------------------------------------------------------------

%-------------------------------------------------------------------------------------
%------------------------------------------------------------------------------

%------------------------------------------------------------------------------
\begin{table}
\small
\centering
\caption{Testing accuracy results}

\setlength\tabcolsep{5pt}
\begin{tabular}{lccc}
\toprule
{\Th{Method}}                  &                    &  &       \\

                    & backbone              &  accuracy$\%$               & epoch           \\ \midrule
                                       \mc{4}{\Th{CIFAR10}}  \\ \midrule

%\midrule

Original                         &LeNet-5&  49.26     & 45    \\
\ee                & LeNet-5   & \tb{54.96} & \tb{33}\\ \midrule

Original                         &CNN-9&  72.61     & 43     \\
\ee                & CNN-9   & \tb{76.64} & \tb{25}\\

\midrule
                                       \mc{4}{\Th{CALTECH101}}  \\ \midrule
       Original       & ResNet-18 &72.83 & 86  \\
       \ee            &  ResNet-18    & \tb{74.39}  & \tb{56}         \\ 
\bottomrule
\end{tabular}
\vspace{0pt}

\label{tab:soa-balanced}
\end{table}
%------------------------------------------------------------------------------

%------------------------------------------------------------------------------
\begin{table}
\small
\centering
\caption{CIFAR10 transformation testing accuracy results}

\setlength\tabcolsep{5pt}
\begin{tabular}{lccc}
\toprule
{\Th{Method}}                  &                    &  &       \\

                    & backbone              &  accuracy$\%$              & epoch           \\ \midrule
                                       \mc{4}{\Th{CIFAR10}}  \\ \midrule

%\midrule
Original + F                        &LeNet-5 &  50.67     & 45    \\
\ee + F                & LeNet-5   & \tb{54.64} & \tb{32}\\ \midrule
Original + C                       &LeNet-5 &  46.74     & 29     \\

\ee + C               & LeNet-5   & \tb{49.90} & \tb{17}\\ \midrule
Original + F+C                       &LeNet-5 &  44.59     & \tb{16}     \\

\ee + F+C             & LeNet-5   & \tb{48.54} & 49\\

\bottomrule
\end{tabular}
\vspace{0pt}

\label{tab:two}
\end{table}
%------------------------------------------------------------------------------
%------------------------------------------------------------------------------
\begin{table}
\small
\centering
\caption{CALTECH101 transformation testing accuracy results}

\setlength\tabcolsep{5pt}
\begin{tabular}{lccc}
\toprule
{\Th{Method}}                  &                    &  &       \\

                    & backbone              &  accuracy$\%$              & epoch           \\ \midrule
                                       \mc{4}{\Th{CALTECH101}}  \\ \midrule

%\midrule
Original + F                        &ResNet-18 &  71.00     & 80    \\ 
\ee + F               & ResNet-18   & \tb{71.96} & \tb{53}\\ \midrule
Original + C                        &ResNet-18 &  79.23     & 88     \\ 
\ee + C              & ResNet-18   & \tb{81.20} & \tb{58}\\ \midrule
Original + C+F                       &ResNet-18 &  78.40     & \tb{89}     \\
\ee + F+C             & ResNet-18   & \tb{79.81} & 92 \\

\bottomrule
\end{tabular}
\vspace{0pt}

\label{tab:three}
\end{table}
%------------------------------------------------------------------------------

\section{\uppercase{conclusion}}
\label{sec:conclusion}
In our research, we introduced a novel approach for enhancing image classification accuracy and computational efficiency. Our method, based on the high boost filtering principle, focuses on edge enhancement. This technique leverages the edge information from labeled data to transform the original dataset, thereby increasing both the size and diversity of the training samples. Through extensive experiments on popular datasets such as CIFAR-10 and CALTECH-101, using well-known neural network architectures including LeNet-5, CNN-9, and ResNet-18, we have demonstrated the effectiveness of our proposed method. Our results substantiate the validity of our hypothesis. Regarding our future research directions, we are interested in exploring the impact of data augmentation techniques in conjunction with our current state-of-the-art approach; additionally, scaling our method to larger datasets such as ImageNet and other domains as semi-supervised learning would be another potential research direction. Our goal is to further enhance classification accuracy while at the same time improving the training efficiency, pushing the boundaries of what is achievable in image classification.

\section*{\uppercase{Acknowledgements}}
In the loving memory of TianHao Bu's father BingXin Bu: 10/01/1965 - 13/09/2023.

\bibliographystyle{apalike}
{\small
\bibliography{reference}}

\end{document}